\documentclass[10pt,twocolumn,letterpaper]{article}
\usepackage{cvpr}              %

\usepackage{graphicx}
\usepackage{amsmath}
\usepackage{amssymb}
\usepackage{booktabs}

\usepackage{xcolor}         %
\definecolor{citecolor}{HTML}{0071bc}
\usepackage[pagebackref,breaklinks=true,letterpaper=true,colorlinks,citecolor=teal,bookmarks=false]{hyperref}
\usepackage{MnSymbol}

\usepackage[capitalize]{cleveref}
\crefname{section}{Sec.}{Secs.}
\Crefname{section}{Section}{Sections}
\Crefname{table}{Table}{Tables}
\crefname{table}{Tab.}{Tabs.}
\newcommand{\refsec}[1]{Section~\ref{sec:#1}}

\newcommand{\reftbl}[1]{Table~\ref{tbl:#1}}
\newcommand{\reffig}[1]{Figure~\ref{fig:#1}}

\newcommand{\refeq}[1]{Equation~\ref{eq:#1}}

\newcommand{\lblfig}[1]{\label{fig:#1}}
\newcommand{\lblsec}[1]{\label{sec:#1}}
\newcommand{\lbleq}[1]{\label{eq:#1}}

\newcommand{\lbltbl}[1]{\label{tbl:#1}}

\definecolor{red3}{HTML}{a40000}
\definecolor{blue3}{HTML}{204a87}
\definecolor{orange2}{HTML}{c17d11}
\definecolor{green3}{HTML}{4e9a06}
\definecolor{gray}{HTML}{babdb6}

\newcommand{\captionvspace}{-3mm}
\newcommand{\tablevspace}{-4mm}
\newcommand{\figvspace}{\tablevspace}

\newcommand{\oursample}{object-guided token sampling\xspace}
\newcommand{\ourattention}{object-aware attention module\xspace}
\newcommand{\ogs}{OGS\xspace}
\newcommand{\oam}{OAM\xspace}
\newcommand{\oaatt}{\oam}
\newcommand{\objectvivit}{ObjectViViT\xspace}

\def\cvprPaperID{****} %
\def\confName{CVPR}
\def\confYear{2023}

\begin{document}

\title{How can objects help action recognition?}

\author{Xingyi Zhou \quad Anurag Arnab \quad Chen Sun \quad Cordelia Schmid\\
Google Research\\
{\tt\small \{zhouxy, aarnab, chensun, cordelias\}@google.com}
}
\maketitle

\begin{abstract}

Current state-of-the-art video models process a video clip as a long sequence of spatio-temporal tokens.
However, they do not explicitly model objects, their interactions across the video, and 
instead process all the tokens in the video.
In this paper, we investigate how we can use knowledge of objects to design better video models, namely to process fewer tokens and to improve recognition accuracy.
This is in contrast to prior works which either drop tokens at the cost of accuracy, or increase accuracy whilst also increasing the computation required.
First, we propose an object-guided token sampling strategy that enables us to retain a small fraction of the input tokens with minimal impact on accuracy.
And second, we propose an object-aware attention module that enriches our feature representation with object information and improves overall accuracy.
Our resulting model, \objectvivit, achieves better performance when using fewer tokens than strong baselines.
In particular, we match our baseline with $30\%$, $40\%$, and $60\%$ of the input tokens on SomethingElse, Something-something v2, and Epic-Kitchens, respectively.
When we use \objectvivit to process the same number of tokens as our baseline, we improve by $0.6$ to $4.2$ points on these datasets.

\end{abstract}

\section{Introduction}
\label{sec:intro}

Video understanding is a central task of computer vision
and great progresses have been made recently with transformer-based models which interpret a video as a sequence of spatio-temporal tokens~\cite{feichtenhofer2019slowfast,arnab2021vivit,bertasius2021space,fan2021multiscale,liu2021video,yan2022multiview}.
However, videos contain a large amount of redundancy~\cite{wu2018compressed}, especially when there is little motion or when backgrounds remain static.
Processing videos with all these tokens is both inefficient and distracting.
As objects conduct motions and actions~\cite{gupta2007objects,wang2018videos}, they present an opportunity to form a more compact representation of a video, and inspire us to study how we can use them to understand videos more accurately and efficiently.
Our intuition is also supported biologically that we humans perceive the world by concentrating our focus on key regions in the scene~\cite{tenenbaum2011grow,grill2005visual}.

\begin{figure}[t]
    \centering
    \includegraphics[width=\linewidth]{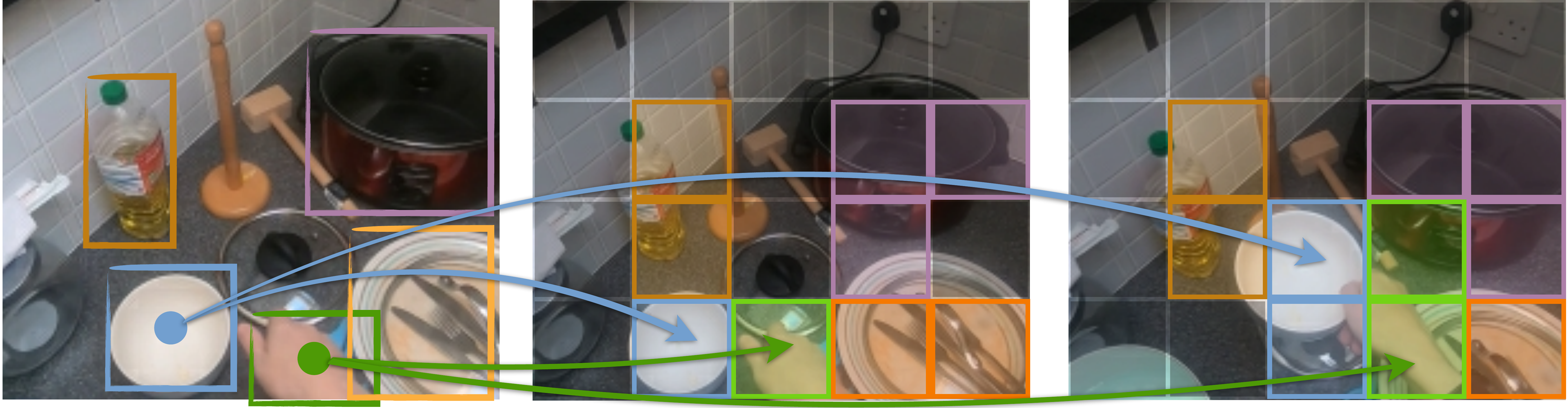}
    \vspace{\captionvspace}
    \caption{
    \textbf{How can objects help action recognition?}
    Consider this action of picking up a bowl on a countertop packed with kitchenware. Objects provide information to: (1) associate image patches (colorful) from the same instance, and identify candidates for interactions; (2) selectively build contextual information from the redundant background patches (dark).
    }
    \vspace{\figvspace}
    \lblfig{new_teaser}
\end{figure}

In this paper, we explore how we can use external object information in videos to improve recognition accuracy, and to reduce
redundancy in the input (\reffig{new_teaser}).
Current approaches in the literature have proposed object-based models which utilize external object detections to improve action recognition accuracy~\cite{herzig2022object,radevski2021revisiting}.
However, they aim to build architectures to model objects and overall bring a notable computational overhead.
We show besides gaining accuracy, objects are extremely useful to reduce token redundancy, and give a more compact video representation.
Fewer tokens also enable stronger test strategies (e.g., multi-crop, longer videos), and overall improve accuracy further.
On the other hand, prior work on adaptive transformers~\cite{rao2021dynamicvit,wang2021efficient,wang2021adaptive,wang2022adafocusv3} dynamically reduce the number of tokens processed by the network conditioned on the input.
Since these methods learn token-dropping policies end-to-end without external hints,
there can be a chicken-and-egg problem: we need good features to know which token to drop, and need many tokens to learn good features.
As a result, they usually suffer from a performance decrease when dropping tokens.
We show we can perform both goals, namely reducing tokens processed by the transformer, as well as improving accuracy, within a unified framework.

\begin{figure*}[t]
    \centering
    \includegraphics[width=0.75\linewidth]{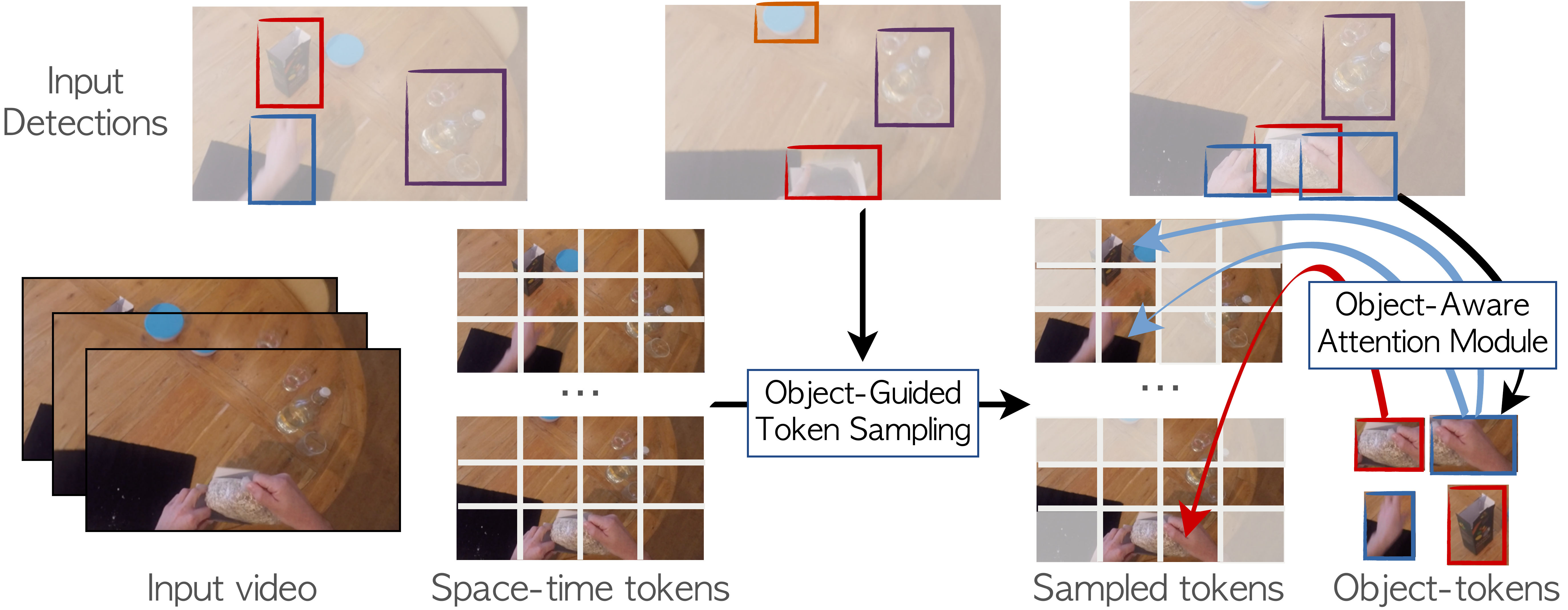}
    \vspace{\captionvspace}
    \caption{
    \textbf{Illustration of our object-based video vision transformer, \objectvivit.}
    \objectvivit takes raw video pixels and off-the-shelf object detections (bounding boxes) as input, and runs space-time attention on video tokens~\cite{arnab2021vivit}.
    We use the detection boxes in two ways.
    (1): we use object locations to downsample the patch tokens before running transformers (Object-Guided Token Sampling, more details in ~\reffig{ogs}).
    (2): we run a customized attention module that creates object tokens from object-patch relations and uses them to enhance patch features (Object-Aware Attention Module, more details in ~\reffig{oaatt}).
    }
    \vspace{\figvspace}
    \lblfig{framework}
\end{figure*}

Concretely, we propose an \textbf{object}-based \textbf{vi}deo \textbf{vi}sion \textbf{t}ransformer, \objectvivit.
As shown in ~\reffig{framework}, we first propose an object-guided token sampling strategy (\ogs) that uses object locations to identify foreground and background tokens.
We retain relevant foreground tokens as is, and aggressively downsample background tokens before forwarding them to the transformer module.
Secondly, to fully leverage the relation between objects and the unstructured spatial-temporal patches, we introduce an object-aware attention module (\oam).
This attention module first creates object tokens by grouping patch tokens from the same object using an object-weighted pooling,
and then applies space-time attention on the concatenated object and patch tokens.
This way, patch features are augmented with their related object information.
Both OGS and OAM are complementary.
They can be used individually to improve either token-compactness or accuracy, or can be used together to get benefits of both.

We validate our method with extensive experiments on SomethingElse~\cite{materzynska2020something}, Something-Something~\cite{goyal2017something}, and the Epic Kitchens datasets~\cite{damen2020epic}.
Using our object-guided token sampling, we find that we can process $60\%\sim90\%$ of the input tokens without losing any accuracy.
And by using our object-aware attention module alone, we outperform a competitive ViViT~\cite{arnab2021vivit} baseline by $0.6$ to $2.1$ points.
Combining both modules, \objectvivit improves token-compactness and accuracy even further,
matching baseline performance by processing $30\%$, $40\%$, and $60\%$ of the input tokens for the three datasets, respectively.
Finally, under the same number of processed tokens but a higher temporal resolution, our model with dropped tokens
improve upon baselines by up to $4.2$ points.
Our code is released at
\href{https://github.com/google-research/scenic/tree/main/scenic/projects/objectvivit}{https://github.com/google-research/scenic}.

\section{Related Work}
\label{sec:related}

\noindent\textbf{Object-based video representation} 
aims to integrate object information in videos~\cite{locatello2020object,shamsian2020learning,elsayed2022savi}
and
is a popular and promising direction in video recognition.
Wang et al.~\cite{wang2018videos} extract RoI features from a 3D CNN feature maps using off-the-shelf detectors, and build a graph neural network on RoI features alone.
LFB~\cite{wu2019long} and Object Transformer~\cite{wu2021towards} load off-the-shelf object features from detectors in video backbones to as feature banks for long video.
Recently, with the advance of transformer-based architectures,
ORViT~\cite{herzig2022object} proposes to crop~\cite{he2017mask} object regions as a new object tokens and attach them to pixel tokens.
ObjectLearner~\cite{zhang2022object} fuses an object-layout stream and a pixels stream using an object-to-pixel transformer.
Most of these existing methods focus on enhancing video features, and so as ours.
However instead of representing objects as cropped boxes, we obtain object features using a simpler heatmap-weighted pool.
In sharp contrast to these methods,
we in addition use objects to decrease transformer token redundancy by using our object-guided token downsampling.

\begin{figure*}[t]
    \centering
    \includegraphics[width=0.9\linewidth]{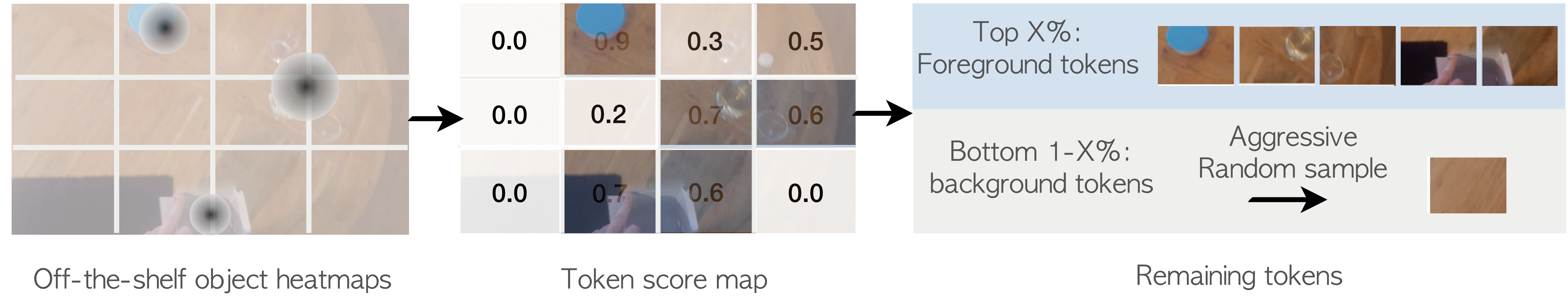}
    \vspace{\captionvspace}
    \caption{
    \textbf{Details about our object-guided token sampling (\textbf{\ogs}).}
    We first render the off-the-shelf object detections as center heatmaps~\cite{zhou2019objects}.
    After tokenization, we define the ``object-scores'' of each token as the sum of the heatmap values inside it.
    We define the tokens with top $X\%$ scores as foreground, where $X$ is a configurable parameter, and the rest as background.
    We keep all foreground tokens, and aggressively downsample the background tokens.
    This mechanism automatically prioritizes object coverage as all object centers have the highest score.
    }
    \lblfig{ogs}
    \vspace{\figvspace}
\end{figure*}

On a parallel direction, STIN~\cite{materzynska2020something} and STLT~\cite{radevski2021revisiting} propose object-layout only architectures without any visual inputs.
These methods only work well with oracle object detections where motions can be perfectly encoded in objects.
However both still report a significant performance drop compared to the pixel-based counterparts.
Our work encodes object information inside a visual backbone by both our sampler and attention module, and retains full-pixel performance with very few visual tokens.

\noindent\textbf{Transformers for videos.}
Transformers~\cite{dosovitskiy2020image} are now the predominant architectures for video recognition.
The vanilla vision transformer (ViT)~\cite{dosovitskiy2020image} evenly divides images into non-overlapping tokens, and runs multi-head self-attention~\cite{vaswani2017attention} over the tokens. 
TimeSFormer~\cite{bertasius2021space} and ViViT~\cite{arnab2021vivit} extends ViT to videos, by introducing tube tokenization and efficient cross-time attention, i.e., axis-based space-time attention~\cite{bertasius2021space} or factorized attention~\cite{arnab2021vivit}.
MotionFormer~\cite{patrick2021keeping} enhances space-time attention by an implicit trajectory attention module.
MViT~\cite{fan2021multiscale,li2021improved} and VideoSwin Transformer~\cite{liu2021video} re-introduce resolution-pooling as in convolution nets in video transformers for efficiency.

Our model is built on top of vanilla space-time attention in ViViT~\cite{arnab2021vivit} without factorization,
which can directly benefit from Masked pretraining~\cite{tong2022videomae} and gives state-of-the-art results~\cite{tong2022videomae,feichtenhofer2022masked}.
We introduce objects as additional sources to improve both token-efficiency and accuracy.

\noindent\textbf{Token-efficiency in transformers.}
Dropping tokens is a popular idea to make transformer models efficient~\cite{rao2021dynamicvit,meng2022adavit,kong2022spvit,fayyaz2021ats}.
DynamicViT~\cite{rao2021dynamicvit} learns a token scoring function end-to-end using Gumbel-Softmax~\cite{jang2016categorical}.
A-ViT~\cite{yin2022vit} adds a loss for the number of active tokens and encourages classifying images with fewer tokens.
ToMe~\cite{bolya2022token} merges tokens by measuring their similarity in test time.
In videos,
STTS~\cite{wang2021efficient} applies a Top-K operator first on time to select keyframes and then on manually designed spatial anchors to select key regions.
AdaFocus~\cite{wang2021adaptive,wang2022adafocus,wang2022adafocusv3}
designs a region selection module using recurrent networks, with an early termination mechanism that stops processing after a certain confidence score.

All these methods learn the token scoring function in an unsupervised way without additional information source.
As a result, they see a reduction in performance as information is reduced.
For example, when STTS~\cite{wang2021efficient} reduces the tokens to $60\%$, accuracy degrades by $1.5\%$.
In contrast, as we use objects to guide our token selection, we do not degrade, and even improve by $0.2\%$ at this dropping ratio.
However efficiency is a nuanced topic~\cite{dehghani2022efficiency}, and whilst our approach does not make the inference pipeline more efficient as it relies on additional object detectors,
it does shine a light on how many tokens we eventually need to process with state-of-the-art transformer architectures.
Furthermore, as we train our model while dropping tokens, it means that we can in fact improve overall training efficiency
over standard video models which do not drop tokens.

\section{Preliminary}
\label{sec:prelim}

Given a video $V$ represented as a stack of pixels $V \in \mathbb{R}^{T \times H \times W \times 3}$,
action classification aims to classify the video into one of the pre-defined action categories $c \in \mathbb{C}$.
We build our framework on top of space-time ViViT~\cite{arnab2021vivit}, which is a straightforward extension of image ViT~\cite{dosovitskiy2020image}.

\textbf{ViViT}~\cite{arnab2021vivit} first divides video pixels $V$ into non-overlapping space-time tokens, where each token represents a pixel tube in size $(dt, dh, dw)$.
This gives $N=\lfloor\frac{T}{dt}\rfloor \times \lfloor\frac{H}{dh}\rfloor \times \lfloor\frac{W}{dw}\rfloor$ tokens.
The RGB pixels inside each token are then mapped to a fixed dimension $D$ by a linear layer, followed by adding positional embeddings.
As a result, the input to the transformer is
$\textbf{z}^0 = [z_1, z_2, \cdots, z_N]$,
where $z_i \in \mathbb{R}^D$ is a patch token.
Each transformer layer updates the tokens $\textbf{z}$ with multi-head attention (MHA)~\cite{vaswani2017attention} and an MLP:
\begin{align}
\lbleq{baseatt}
\textbf{y}^l &= \text{MHA}(\textbf{z}^l, \textbf{z}^l, \textbf{z}^l) + \textbf{z}^l \\
\textbf{z}^{l + 1} &= \text{MLP}(\textbf{y}^l) + \textbf{y}^l
\end{align}
Where MHA(q, k, v) computes attentions between queries $q$, keys $k$, and values $v$.
MLP contains $2$ linear layers. We omit layer norm~\cite{ba2016layer} for simplicity.
The output classification logits $\textbf{c} \in \mathbb{R}^{|\mathbb{C}|}$ is obtained after a linear layer on global average-pooled features of all tokens in the last layer.

\section{Method}
\label{sec:method}

Besides video pixels $V$, our framework takes an additional input: object detections $\mathcal{B} = \{\{(b_{t, o}, \delta_{t, o})\}_{o=1}^{n_t}\}_{t=1}^{T}$, where $b_{t, o} = (x_{t, o}, y_{t, o}, w_{t, o}, h_{t, o})$ is the bounding box of the $o$-th object in frame $t$, $\delta_{t, o}$ is an optional object identity (when tracking is enabled), and $n_t$ is the number of objects in frame $t$.
The bounding boxes are usually from an off-the-shelf object detector~\cite{ren2015faster}, or can be oracle boxes from annotation for analytic benchmarks~\cite{materzynska2020something}.
Next, we introduce \objectvivit, which includes two components, Object-guided token sampling (\ogs) and Object-aware attention module (\oaatt), to use these external detections for token-efficient and accurate action recognition.

\subsection{Object-guided token sampling}
\lblsec{sample}
We first use objects $\mathcal{B}$ to downsample input tokens $\textbf{z}^0$.
\reffig{ogs} gives an overview of this process.
Our main insight is that tokens which are inside objects carry the motions in the scene and should be retained, and tokens in the background are mostly static and can be partially dropped due to their redundancy.
We also aim to configure the dropping ratio to be able to control the token-accuracy trade-off.

To do so, we define a continuous token objectness score by measuring how close each token is to object centers.
Specifically, we follow CenterNet~\cite{zhou2019objects} to render objects in each frame into a single class-agnostic center heatmap $\textbf{H}_t \in \mathbb{R}^{H \times W}$ in the original image resolution, where
\begin{equation}
\textbf{H}_{t,i,j} = \max_{o\leq n^t} \exp{\left(-\frac{(i - x_{t, o})^2 + (j - y_{t, o})^2}{2\sigma_{t, o}^2}\right)},
\end{equation}
and $\sigma_{t, o}$ is a monotonic function of the object size that controls the Gaussian radius.
We use the same tube size $(dt, dh, dw)$ to project the heatmap into tubelets $\widetilde{\bf H} \in \mathbb{R}^{N}$.

We simply select the top $X\%$ of the tokens according to object-ness scores $\widetilde{\textbf{H}}$:
\begin{equation}
\textbf{z}^0_{\text{fg}} = \{z_n | \widetilde{\textbf{H}}_n \ge \tau \}, \quad
\textbf{z}^0_{\text{bg}} = \textbf{z}^0 - \textbf{z}^0_{\text{fg}}
\end{equation}
where $\tau$ is the $X\%$-th value of $\widetilde{\textbf{H}}$, and $X$ is a configurable parameter to control the number of tokens.

With this foreground/ background token definition, we keep all the foreground tokens as is, and uniformly sample $Y\% \times N$ tokens from the background.
This gives the downsampled background tokens $\widehat{\textbf{z}}^0_{\text{bg}}$, where $|\widehat{\textbf{z}}^0_{\text{bg}}| = Y\% \times N$.
Our final inputs are a concatenation of both tokens:
\begin{equation}
\widehat{\textbf{z}}^0 = [\textbf{z}^0_{\text{fg}}, \widehat{\textbf{z}}^0_{\text{bg}}]
\lbleq{tokensample}
\end{equation}
We therefore have $|\widehat{\textbf{z}}^0| = (X + Y)\% \times N$ remaining tokens.

\begin{figure}
    \centering
    \includegraphics[width=\linewidth]{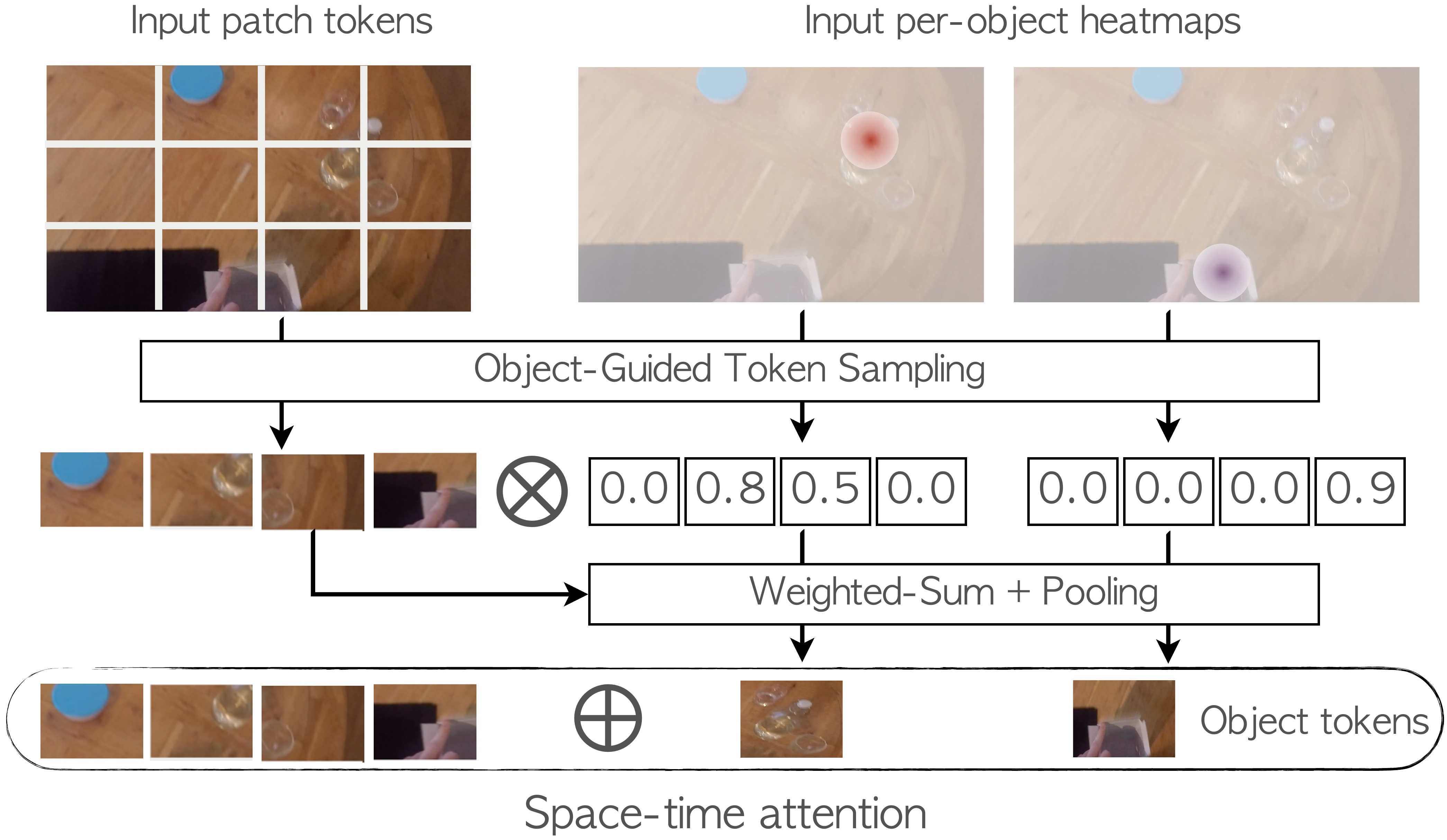}
    \vspace{\captionvspace}
    \caption{
    \textbf{Details about our object-aware attention module (\textbf{\oaatt}).}
    The inputs to the module are (downsampled) space-time patch tokens together with their object-scores with respect to each object instance.
    We only show one frame for clarity.
    For each object instance at each frame, we create a new object token by weighted pool between the pixel features and the object weights ($\bigotimes$).
    We concatenate ($\bigoplus$) the object tokens and patch tokens and conduct space-time attention over all tokens following ORViT~\cite{herzig2022object}.
    The module outputs updated features of patch tokens, which are enhanced by object information.
    }
    \lblfig{oaatt}
    \vspace{\figvspace}
\end{figure}

\subsection{Object-aware attention}
\lblsec{attention}
Next, we use objects $\mathcal{B}$, in transformer blocks to improve features (\reffig{oaatt}).
We again represent objects in a frame as a center-heatmap~\cite{zhou2019objects}, but here we use instance-specific heatmaps.
I.e., for each object $o$ in frame $t$, we render a heatmap $\textbf{H}_{o, t} \in \mathcal{R}^{H \times W}$ with a single Gaussian peak:
$\textbf{H}_{o, t,i,j} = \exp{\left(-\frac{(i - x_{t, o})^2 + (j - y_{t, o})^2}{2\sigma_{t, o}^2}\right)}$.
Each $\textbf{H}_{o, t}$ highlights the pixel regions of object $o$ in frame $t$.
After tokenization and the token-sampling following \refeq{tokensample}, the downsampled per-instance heatmap $\widehat{\textbf{H}}_{o, t}$ represents the affinity score between object $o$ and each remaining token in frame $t$.
These affinity scores naturally serve as weighting scores to aggregate object features, and we directly use them to create object tokens $\textbf{w}_{o, t}$ for each object at each frame.
\begin{equation}
\textbf{w}_{o, t}^l = \text{MaxPool}(\text{MLP}(\widehat{\textbf{H}}_{o, t} * \widehat{\textbf{z}}^l_t)),
\lbleq{objpool}
\end{equation}
where $l$ is the transformer layer index,
$\widehat{\textbf{z}}^l_t$ is the (downsampled) patch tokens of frame $t$ at layer $l$, and MLP are two linear layers following ORViT~\cite{herzig2022object}.
Optionally, when object identity information is available (through tracking), we encode object identity via a learnable identity embedding $\textbf{e}_{o} \in \mathbb{R}^D$ shared across time:
$\textbf{w}_{o, t}^l = \textbf{w}_{o, t}^l + \textbf{e}_o^l.$

We next concatenate the object tokens and the (downsampled) patch tokens and keys and values, and use them to update the query patch tokens in \refeq{baseatt}:
\begin{equation}
\textbf{y}^l = \text{MHA}(\textbf{z}^l, [\textbf{z}^l, \textbf{w}^l], [\textbf{z}^l, \textbf{w}^l]) + \textbf{z}^l
\end{equation}

Following ORViT~\cite{herzig2022object}, we observe inserting the object tokens at 3 layers (the 2, 6, 11-th layer) of the transformer is sufficient. All other layers use the vanilla self-attention in \refeq{baseatt}.

\noindent\textbf{Comparison to ORViT}~\cite{herzig2022object}.
Our object-aware attention module is inspired by the Object-Region Attention of ORViT~\cite{herzig2022object}.
We both create object tokens from pixels and involve them in cross-attention computation.
ORViT uses ROIAlign~\cite{he2017mask} to aggregate object features, which requires the input to be a full rectangular grid.
They also use an axis-aligned trajectory attention backbone~\cite{patrick2021keeping} which requires the same shape across frames.
We use a simpler token aggregating function (\refeq{tokensample}) that does not have these constraints and works for any input shape.
We also don't use the bounding box layout module in ORViT~\cite{herzig2022object}.
We will show our formulation is simpler, more flexible, and performs competitively.

\section{Experiments}
\label{sec:experiments}

\begin{figure*}[t]
    \centering
    \begin{subfigure}{0.33\linewidth}
        \includegraphics[width=\linewidth, page=1]{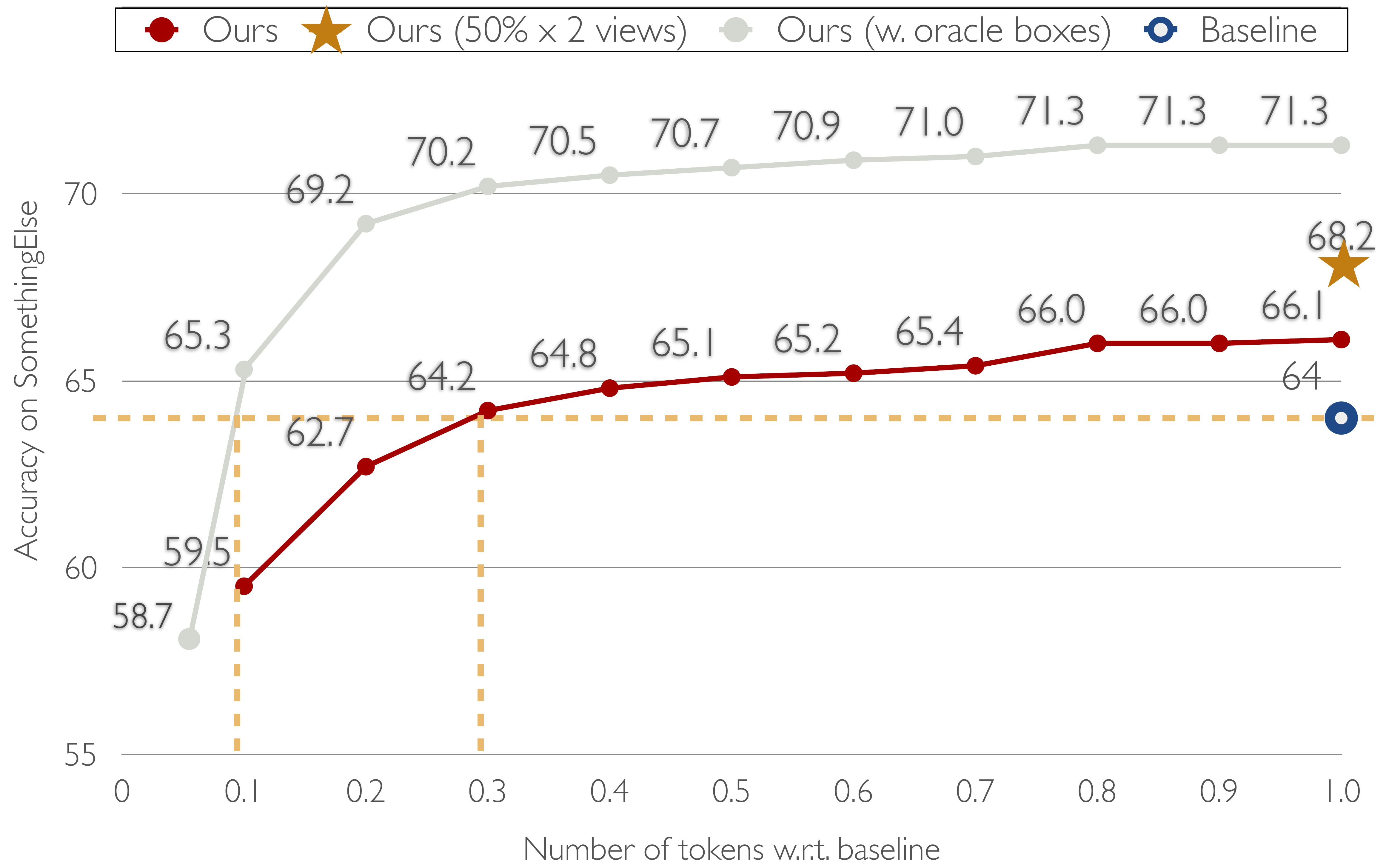}
        \caption{SomethingElse}
        \lblfig{results:else}
    \end{subfigure}
    \begin{subfigure}{0.33\linewidth}
        \includegraphics[width=\linewidth, page=2]{figures/plots_all_v3.2.pdf}
        \caption{Something-Something v2}
        \lblfig{results:ssv2}
    \end{subfigure}
    \begin{subfigure}{0.33\linewidth}
        \includegraphics[width=\linewidth, page=3]{figures/plots_all_v3.2.pdf}
        \caption{Epic-Kitchens}
        \lblfig{results:epic}
    \end{subfigure}
    \vspace{\captionvspace}
    \vspace{-2mm}
    \caption{
    \textbf{Main results.} We show action classification results with our proposed \textbf{Object-aware attention} module and \textbf{Object-guided token sampling} under different numbers of sampled tokens ($(X+Y)\%$, the ratio with respect to all tokens) on SomethingElse (a), SSv2 (b), and Epic Kitchens (c) dataset.
    We show results with inferred boxes from detectors ((\textcolor{red3}{Ours})) or with \textcolor{gray}{oracle boxes}.
    Our baseline uses Space-time ViViT~\cite{arnab2021vivit} with VideoMAE pretraining~\cite{tong2022videomae}.
    The models are evaluated under single-crop and single-view testing by default.
    Using all tokens (x-axis at 1.0), our Object-aware attention improves the baseline by $0.6$ to $2.1$ points.
    With fewer tokens, our model matches the baseline performance with $30\%$, $40\%$, and $60\%$ of tokens on the three datasets, respectively.
    We also show an optimal model (\textcolor{orange2}{$\bigstar$}) with $50\%$ tokens and 2-view testing, which achieves better accuracy than the full model under the same number of processed tokens.
    With ground truth boxes, we only need $10\%$ tokens to match the baseline's performance.
    }
    \vspace{\tablevspace}
    \lblfig{main-results}
\end{figure*}

We evaluate \objectvivit for action classification on Something-something v2~\cite{goyal2017something}, SomethingElse~\cite{materzynska2020something}, and Epic Kitchen~\cite{damen2020epic} datasets.

\noindent\textbf{Something-something v2 (SSv2)}~\cite{goyal2017something} is a large-scale action classification dataset consisting of $\sim\!169$K training videos and $\sim\!25$K validation videos.
The videos are $2$-$4$ seconds long, focusing on hand-object interaction actions, with in total 174 action classes.
The action classes are agnostic to object classes.
\textbf{Something-Else}~\cite{materzynska2020something} is a re-split of SSv2 videos, with a focus on compositional action recognition regardless of object classes.
It ensures object classes of the same action are disjoint between training and validation.
The resulting split has $\sim\!55$K training videos and $\sim\!67$K validation videos.
The videos are decoded in $12$ FPS.

The Something-Else authors additionally annotate ground-truth object bounding boxes with tracked identities for videos in its split.
In our analytic experiments, we use these oracle box annotations to study the performance upper-bound of our framework.
In our main experiments, we use inferred bounding boxes from ORViT~\cite{herzig2022object}, which are from a finetuned FasterRCNN detector~\cite{ren2015faster,wu2019detectron2} with ResNet-101 backcbone on the corresponding splits of the SomethingElse box annotations.

\noindent\textbf{Epic-Kitchens}~\cite{damen2020epic} contains $67$K training videos and $10$K validation video segments on kitchen scenes, where each segment is on average $3$ seconds long.
The action is composed of a verb (from $97$ classes) and a noun (from $300$ classes), and is considered as correct if both verb and noun classification are correct.
The dataset does not provide bounding box annotations, and we follow ORViT~\cite{herzig2022object} to use an off-the-shelf object detector trained on hand-object interaction dataset~\cite{shan2020understanding}, and use a learning-free tracker SORT~\cite{bewley2016simple} to link them.
We decode the videos at $60$ FPS~\cite{herzig2022object}.

\subsection{Implementation details}
We implement our method in JAX~\cite{jax2018github} based on the Scenic library~\cite{dehghani2021scenic}.
We use space-time ViViT~\cite{arnab2021vivit} with VideoMAE pretraining~\cite{tong2022videomae} as our baseline model.
In all experiments, we use input size of $(16, 224, 224)$ and token cube size of $(2, 16, 16)$, which results in an input size of $1568$ tokens.
Our training hyper-parameters follow VideoMAE finetuning~\cite{tong2022videomae}.
Specifically, we train our model and baselines using the AdamW optimizer~\cite{kingma2015adam} with learning rate $10^{-3}$ and batch size of 256 videos with a cosine learning rate decay.  %
During training, we resize images to short side $256$ and do a random crop at $224$.
Unless specified, during testing we resize the image to short side $224$, and do a single center crop.
We use uniform frame sampling following VideoMAE~\cite{tong2022videomae} and TSN~\cite{wang2018temporal}.
I.e., the sampling stride is proportional on the video length so that the sampled frames always cover the whole video.
We use regularizations including drop path~\cite{huang2016deep} of $0.1$, label smoothing~\cite{szegedy2016rethinking} of $0.1$, and mixup~\cite{zhang2017mixup} of $0.8$.
On SSv2, we load VideoMAE checkpoint pretrained on SSv2 training set~\cite{tong2022videomae}, and train our model for 30 epochs.
On SomethingElse and Epic Kitchens, we initialize from the VideoMAE checkpoint pretrained on Kientics400~\cite{kay2017kinetics} publicly released by the authors~\cite{tong2022videomae}, and train our model for 50 epochs.
All other training parameters are the same between datasets.

\begin{figure*}
    \centering
    \begin{subfigure}{0.33\linewidth}
        \includegraphics[width=\linewidth, page=1]{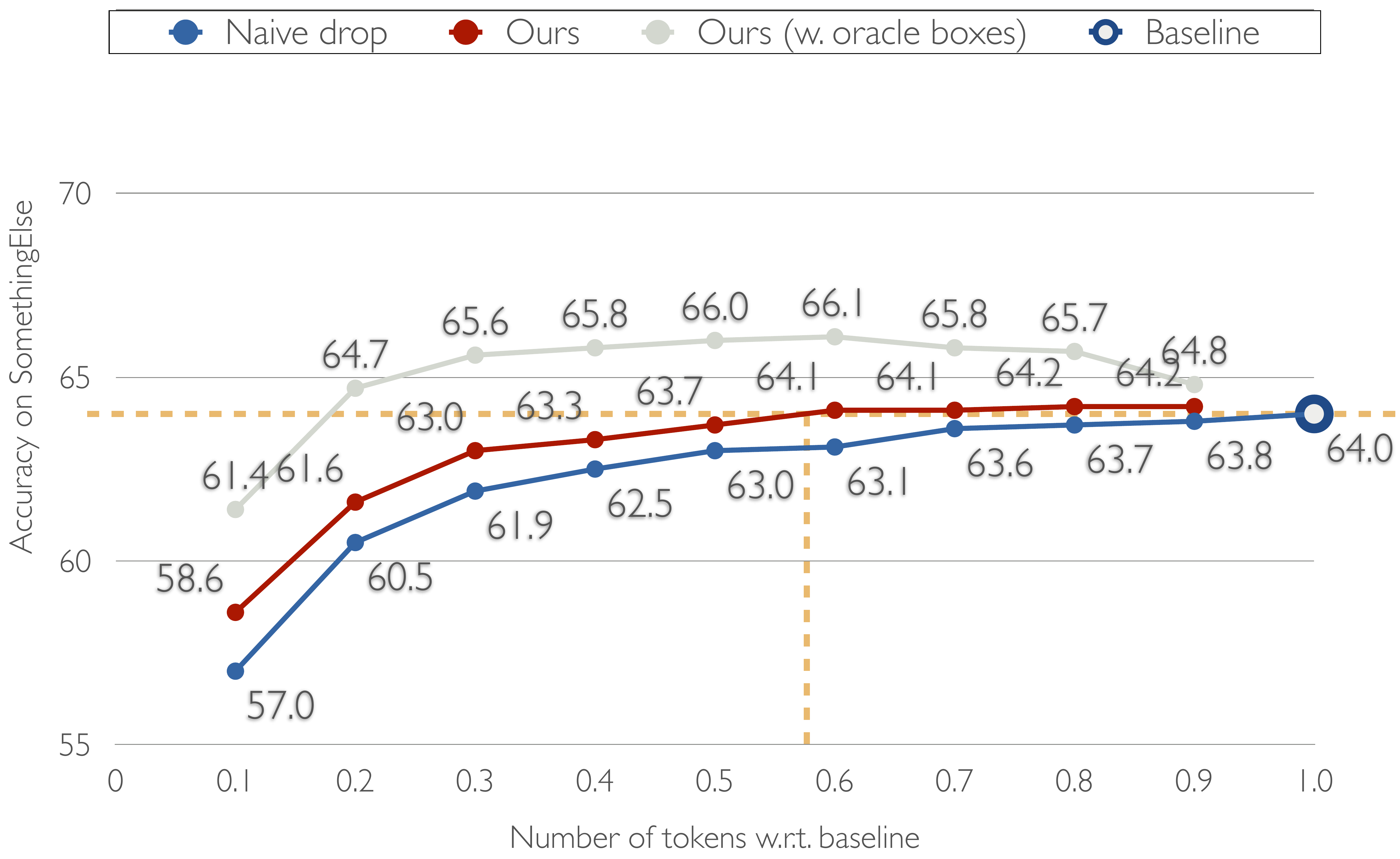}
        \caption{SomethingElse}
        \lblfig{results:else}
    \end{subfigure}
    \begin{subfigure}{0.33\linewidth}
        \includegraphics[width=\linewidth, page=2]{figures/plots_sample_v3.1.pdf}
        \caption{Something-Something v2}
        \lblfig{results:ssv2}
    \end{subfigure}
    \begin{subfigure}{0.33\linewidth}
        \includegraphics[width=\linewidth, page=3]{figures/plots_sample_v3.1.pdf}
        \caption{Epic-Kitchens}
        \lblfig{results:epic}
    \end{subfigure}
    \vspace{\captionvspace}
    \vspace{-1mm}
    \caption{
    \textbf{Effectiveness of Object-guided sampling.} We show action classification accuracy on SomethingElse (a), SSv2 (b), and Epic Kitchens (c) dataset with our \oursample only under different numbers of tokens (without Object-aware attention). We use the same space-time ViViT baseline as \reffig{main-results}. We compare to a naive-drop baseline that uniformly random sample tokens.
    Our object-guided sampling consistently outperforms the naive sample baseline on all sampling ratios and datasets. Surprisingly, we observe that token-sampling alone can improve action-recognition performance on SomethingElse and SSv2 (with both inferred or oracle boxes), and matches the baseline performance with $50\% \sim 60\%$ number of tokens.
    }
    \lblfig{sample}
\vspace{-2mm}
\end{figure*}

\subsection{Main results}

We first show our full model results with both our proposed modules, \oursample (\ogs) and \ourattention (\oam).
The hyper-parameters of our model include the foreground token split ratio $X\%$ and the background sampling ratio $Y\%$ (\refsec{sample}).
We keep the background sampling ratio $Y\%$ = $10\%$ (as will be ablated in \reffig{sample-bg}, except for $10\%$ total tokens where we use $Y\%$ = $5\%$), and sweep $X\%$ so that the total number of remaining tokens $(X+Y)\%$
range from $10\%$ to $90\%$.
For the results of $100\%$ tokens, we do not apply \ogs and only use \oam.

~\reffig{main-results} shows the results on SomethingElse, SSv2, and Epic-Kitchens under different number of tokens.
First, we observe that without dropping tokens (x-axis at 1.0), \oam effectively boosts the space-time ViViT baseline~\cite{arnab2021vivit} by a healthy margin, with $2.1\%$, $1.3\%$, and $0.6\%$ points improvements on the three datasets, respectively.
For analysis purposes, we also show results with the ground truth detections (in both training and testing) when available.
Here the improvements further increase to $7.3\%$ and $6.3\%$ on SomethingElse and SSv2, respectively.
This implies our performance can be further improved with stronger detectors.

When starting dropping tokens using \ogs, the performance of our model decreases due to dropped input information.
The performance drops mildly with respect to tokens, and overall yields a favourable trade-off.
Specifically, on SomethingElse, SSv2, and Epic-Kitchens, our model meets the baseline's performance at $30\%$, $40\%$, and $60\%$ tokens, respectively.
This convincingly shows object regions are right highlights of the videos and can effectively guide token selection.
Again for analysis purposes, we show when ground truth bounding boxes are available, we can match the baseline performance with only $10\%$ tokens.

As our model saves tokens in inference, we configure a testing strategy that applies our $50\%$-token model with 2-view testing (i.e., uniform sampling frames with different starting frames and averaging output logits)
so that the
overall number of tokens processed matches the baseline.
~\reffig{main-results} shows this model (\textcolor{orange2}{$\bigstar$}) further improves our full-token model.
This again highlights that tokens-efficiency is an important aspect in action recognition.

\begin{figure}[t]
\centering
\includegraphics[width=0.95\linewidth]{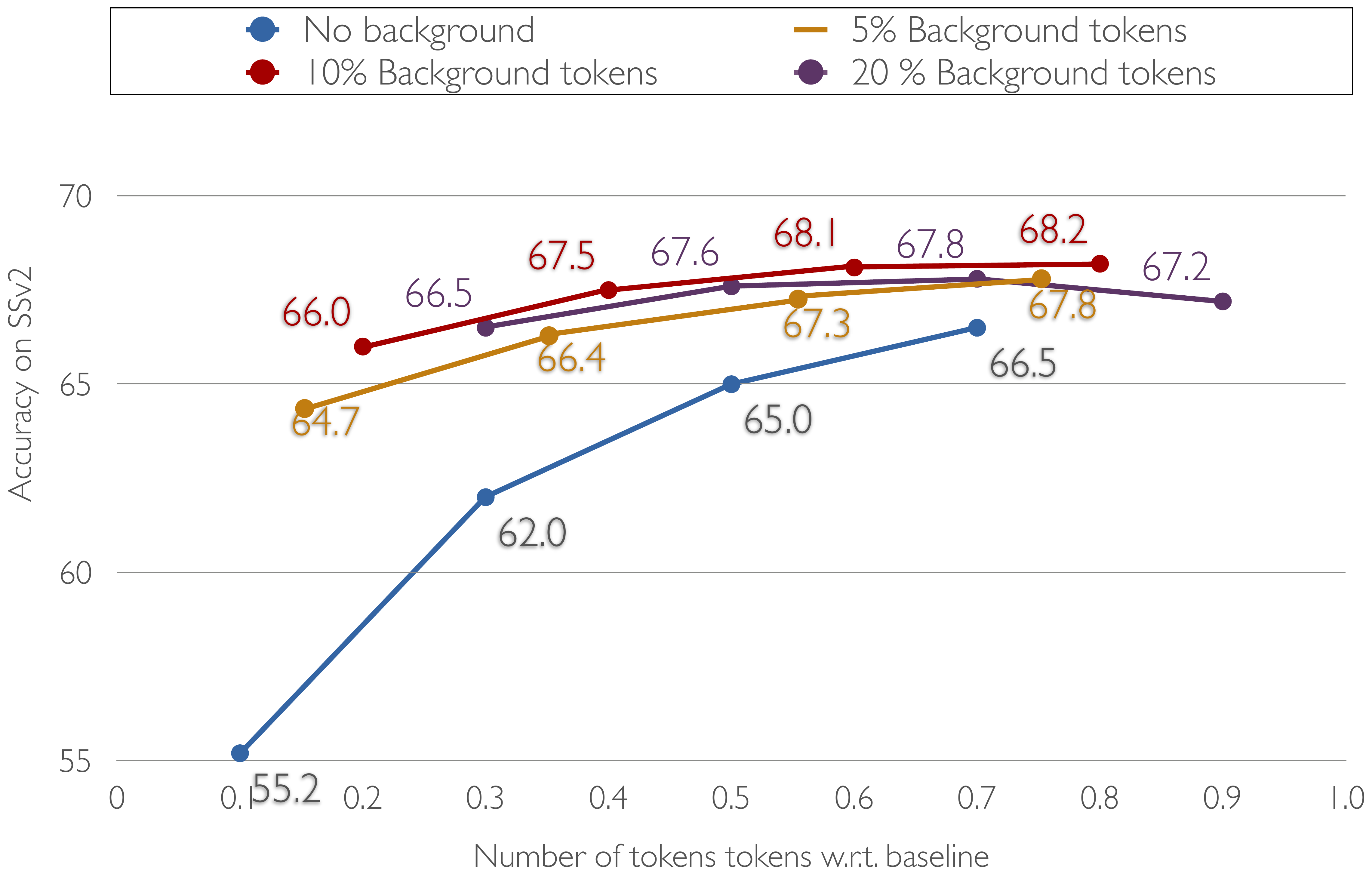}
\vspace{\captionvspace}
\caption{
\textbf{Importance of adding background tokens in \oursample.}
We show action recognition accuracy under different numbers of remaining tokens ($(X+Y)\%$ in x-axis) with different amounts of background tokens (different $Y\%$ in different lines).
We experiment using ground truth boxes on Something-something V2 dataset with \ogs.
We observe the model without any background tokens performs poorly, and $10\%$ background tokens give the best overall performance.
}
\vspace{-5mm}
\lblfig{sample-bg}
\end{figure}

\begin{table*}[t]
\centering
\begin{subfigure}{0.31\linewidth}
\centering
\begin{tabular}{@{}l@{\ \ \ \ \ \ }c@{\ \ }c@{}}
\toprule
Layer index & Top 1 Acc. & gFLOPs \\
\midrule
*0 & 66.1 & 80\\
1 & 66.0 & 88\\
6 & 65.9 & 130\\
\bottomrule
\end{tabular}
\caption{
\textbf{When to drop tokens.}
We apply our sampler on different layers (without object attention), with $10\%$ background tokens and $40\%$ foreground tokens.
}
\lbltbl{whentodrop}
\end{subfigure}
\ \ 
\begin{subfigure}{0.31\linewidth}
\centering
\begin{tabular}{@{}l@{\ \ }c@{}}
\toprule
 & Top 1 Acc. \\
\midrule
RoIAlign & 67.2 \\
Binary block mask & 67.2 \\
*Center heatmap weighting & 67.4 \\
\bottomrule
\end{tabular}
\caption{
\textbf{Object feature aggregation functions.}
We compare RoIAlign used in ORViT~\cite{herzig2022object}, a simpler block masking, and our proposed heatmap weighting.
}
\lbltbl{aggregation}
\end{subfigure}
\ \ 
\begin{subfigure}{0.31\linewidth}
\centering
\begin{tabular}{@{}l@{\ \ }c@{}}
\toprule
 & Top 1 Acc. \\
\midrule
w.o. tracking & 67.0 \\
w. identity attention & 67.4 \\
*w. identity embedding & 67.4 \\
\bottomrule
\end{tabular}
\caption{
\textbf{How to use tracking information.}
We compare no tracking, using additional layers for tokens for the same object, or simply adding an identity embedding.
}
\lbltbl{track}
\end{subfigure}
\vspace{\captionvspace}
\caption{
\textbf{Ablation studies.} We ablate design choices for both \oursample (\ref{tbl:whentodrop}), and \ourattention (\ref{tbl:aggregation}, \ref{tbl:track}). All models use inferred boxes on SSv2~\cite{goyal2017something}. * means our default option.
}
\lbltbl{ablation}
\vspace{\figvspace}
\end{table*}

\vspace{1mm}
\noindent\textbf{Effectiveness of Object-guided sampling.}
Next, we apply only \ogs without \oam to study the importance of the sampler alone.
Following the setting in \reffig{main-results}, we vary the foreground token split ratio $X\%$ and show the action classification change of the three datasets.
We compare to a naive token dropping baseline that uniform-randomly drop tokens in space and time, which is equivalent to setting $X\%=0$ and varying $Y\%$ in our framework, and is the same as the token sampler used in video MAE pretraining~\cite{feichtenhofer2022masked}.
\reffig{sample} shows the results.
On all the three datasets and all sampling ratios, our object-guided sampler outperforms the uniform sampling baseline, with a more significant gain when the total sampling ratio is low.
Surprisingly, we observe solely dropping tokens can improve action classification accuracy for free: on all the three datasets, we observe a $0.2$ points gain on their corresponding optimal sampling ratio.
This is likely thanks to dropping background tokens highlights the foreground motions and makes the learning task easier.
This phenomenon is more pronounced when we use the oracle detections, where we observed a $2.1\%$ gain on both Something datasets.
The performance improvements do not increase with more tokens, as $100\%$ tokens go back to the no-sampling baseline.
Most importantly, on all the three datasets, we all match the full-token baseline with fewer tokens, with $60\%$, $50\%$, $90\%$ tokens, respectively.
\reffig{qual} shows qualitative results of our sampler.

\subsection{Ablation studies}

Next, we ablate design choices in both our proposed modules, including the importance of background tokens, when to apply token sampling, and designs in the object-aware attention module.

\noindent{\textbf{Importance of adding background tokens}}.
We first show it is crucial to include both objects and contexts.
\reffig{sample-bg} shows the accuracy trade-off under different amounts of background tokens on SomethingSomeghing-V2 (with ground truth bounding boxes).
The entry without any background tokens significantly underperforms other entries even with very few background tokens ($5\%$).
This is because non-object tokens provide important contextual information that might be critical for the object, e.g., showing the action is happening on a desk or a dining table.
Background tokens also prevent the model from overfitting into inaccurate object detection or annotation.
Our experiments show including $10\%$ background tokens gives an overall good trade-off, but different foreground-background token split ratios do not crucially impact the results.

\noindent\textbf{When to drop tokens.}
\ogs takes the full set of token as input and produce downsampled tokens.
It can be applied to any blocks in the transformer. 
Intuitively, applying token downsampling in later layers~\cite{ryoo2021tokenlearner} reserves more information with
the cost of more computation.
\reftbl{whentodrop} ablates doing token sampling in different layers.
We observe dropping tokens at the very beginning before any transformer layers work as well as dropping in middle blocks,
while being the most computationally efficient.
We thus use this early dropping by default.

\noindent\textbf{Object feature aggregation functions.}
\oam aggregates patch features into object features.
A popular way~\cite{wang2018temporal,wu2019long,herzig2022object} to do that is to crop and resize (i.e., RoIAlign~\cite{he2017mask}) the box region in the feature grids.
However, this is no longer feasible when the input features are unstructured patches after downsampling.
In \reftbl{aggregation}, we compare alternatives to RoIAlign under the full token settings.
We first use a binary block mask that masks out input features outside of the bounding box region, followed by linear layers and a global pooling.
This matches RoIAlign without cropping.
Our heatmap-weighted pooling (\refeq{objpool}) further improves the block-mask, likely due to the center heatmap giving a better estimation of the object mask.
A mask-guided feature aggregation~\cite{cheng2022masked} might perform better, when instance segmentation annotations are available.

\begin{table}[b]
\vspace{\tablevspace}
\centering
\small
\begin{tabular}{@{}l@{\ }c@{\ \ \ }c@{\ \ }c@{}}
\toprule
& SSv2 & S.Else & n. FLOPs\\
\midrule
MFormer~\cite{patrick2021keeping} & 66.5 & 62.8  & $1\times$ (370) \\
MFormer w. ORViT~\cite{herzig2022object} & 67.9 / - & 69.7  & $1.09\times$ \\
MFormer w. ObjectLearner~\cite{zhang2022object} & - / 74.0 & 73.6  & $1.04\times$ \\
\midrule
VideoMAE (our baseline) & 64.8 & 67.0& $1\times$ (181)\\
VideoMAE w. ORViT (reproduced)& 68.3 / 72.8 & 71.6 & $1.06\times$\\
\objectvivit & 68.5 / 73.0 & 72.0  & $1.06\times$\\ 
\objectvivit ($50\%$ tokens $\times 2$ views) & 70.0 / 74.2 & 74.5  & $1.06\times$\\ 
\bottomrule
\end{tabular}
\vspace{\captionvspace}
\caption{
\textbf{
Compare to existing object-based video models~\cite{herzig2022object,zhang2022object}.
}
We report normalized FLOPs (n. FLOPs) with respect to the corresponding baseline.
We show results on SomethingElse (S.Else) with {ground truth} boxes and and SSv2 using both {inferred} and ground truth boxes (results shown in format: inferred boxes / ground truth boxes).
The results are reported with a single temporal view and 3 spatial crops~\cite{herzig2022object} except for the last row,
which uses 3 spatial crops and 2 temporal views with $50\%$ tokens.
}
\lbltbl{sota-comparison}
\end{table}

\begin{figure*}[t]
\centering
\begin{tabular}{@{}c@{}c@{\ }c@{\ }c@{\ \ }c@{}c@{\ }c@{\ }c@{}}
RGB & Heatmap & FG-tokens & Sampled-tokens & RGB & Heatmap & FG-tokens & Sampled-tokens \\
\includegraphics[width=.24\columnwidth]{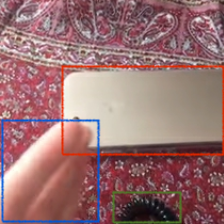} &
\includegraphics[width=.24\columnwidth]{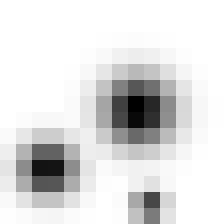} &
\includegraphics[width=.24\columnwidth]{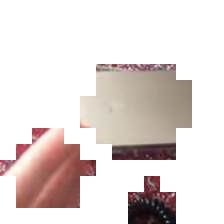} &
\includegraphics[width=.24\columnwidth]{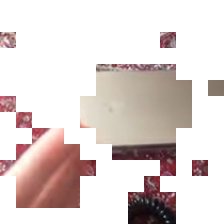} &
\includegraphics[width=.24\columnwidth]{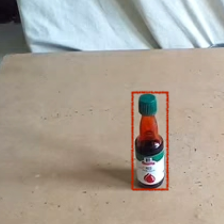} &
\includegraphics[width=.24\columnwidth]{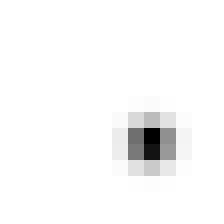} &
\includegraphics[width=.24\columnwidth]{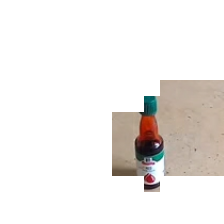} &
\includegraphics[width=.24\columnwidth]{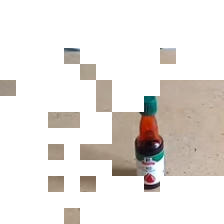} \\
\includegraphics[width=.24\columnwidth]{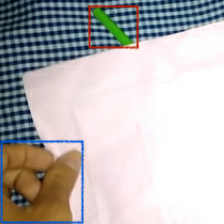} &
\includegraphics[width=.24\columnwidth]{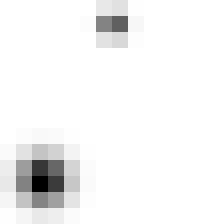} &
\includegraphics[width=.24\columnwidth]{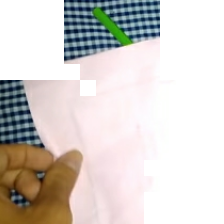} &
\includegraphics[width=.24\columnwidth]{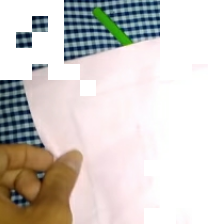} &
\includegraphics[width=.24\columnwidth]{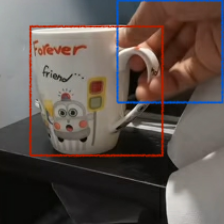} &
\includegraphics[width=.24\columnwidth]{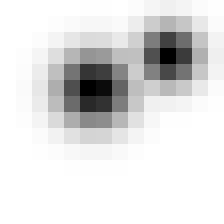} &
\includegraphics[width=.24\columnwidth]{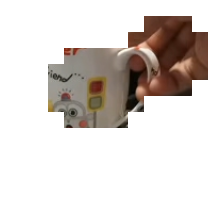} &
\includegraphics[width=.24\columnwidth]{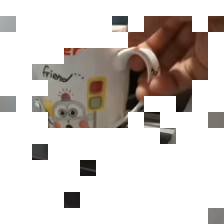} \\
\end{tabular}
\vspace{\captionvspace}
\caption{{\textbf{Qualitative results of our \oursample.}
A row shows two frame from two different videos.
From left to right, we show the original RGB pixels with external bounding boxes, the object-heatmap used to label tokens, the resulting foreground tokens, and the overall sampled tokens including background.
We use foreground split ratio $X=30\%$ and background sample ratio $Y=10\%$.
}}
\vspace{\tablevspace}
\vspace{-1mm}
\lblfig{qual}
\end{figure*}

\noindent\textbf{How to use tracking information.}
By default, our \ourattention does not use object tracking information as each object token is created per frame individually.
We study options to add identity information in our framework when available (e.g., via a light-weighted tracker~\cite{bewley2016simple}).
\reftbl{track} shows the results.
Without any tracking information, our \ourattention still improves the baseline ($66.1$) by $0.9$ points.
We experimented with two ways to use tracking.
The first is to apply an additional attention module on object tokens from the same track,
and the second is to simply add the same embedding to objects within a track following~\cite{herzig2022object}.
We observe both further improve the performance by $0.4$ points, and we choose to use the embedding-based way for simplicity.

\subsection{Comparisons to existing work}

Next, we compare to existing methods on both Object-based video modeling and token-efficient transformers.

\noindent\textbf{Object-based video representation.}
We compare our full model to ORViT~\cite{herzig2022object} and ObjectLearner~\cite{zhang2022object} on SomethingElse~\cite{materzynska2020something} and SSv2 datasets~\cite{goyal2017something}.
Following their evaluation protocol, we use ground truth boxes on SomethingElse and the same inferred boxes on SSv2.
Both methods use MotionFormer~\cite{patrick2021keeping} as their baseline which requires full grid-shaped inputs, which are different from our vanilla ViT baseline.
Thus, we reproduce ORViT~\cite{herzig2022object} on our baseline based on their released code.
We report their original numbers quoted from their publications as well as our reproduced numbers on our baseline.
\reftbl{sota-comparison} show our full model slightly outperform ORViT~\cite{herzig2022object}, mostly thanks to our better object feature aggregation functions.
We highlight our additional ability to process with tokens.
When using our optimal model with $50\%$ tokens and multi-view testing, our gains are larger under the same normalized FLOPs.

\begin{table}[b]
\centering
\vspace{-2mm}
\vspace{\tablevspace}
\begin{tabular}{@{}l@{\ }c@{\ \ }c@{\ }c@{}c@{}}
\toprule
 & \!\!Ratio & Baseline & Proposed & $\Delta$\\
\midrule
End-to-end \\
\ \ STTS~\cite{wang2021efficient} & $70\%$ & 69.6~\cite{liu2021video} & 68.7 & $-0.9\%$ \\
\ \ AdaFocus v3~\cite{wang2022adafocusv3} & $44\%$ & 59.6~\cite{lin2019tsm} & 59.6 & $+0.0\%$\\
\midrule
Object-guided \\
\ \ {\objectvivit} & {$40\%$} & {66.1}~\cite{tong2022videomae} & {66.2} &{$+0.1\%$} \\ 
\ \ \textcolor{gray}{\objectvivit w. gt boxes} & \textcolor{gray}{$15\%$} & \textcolor{gray}{66.1}~\cite{tong2022videomae} & \textcolor{gray}{67.4} & \textcolor{gray}{$+1.3\%$} \\ 
\bottomrule
\end{tabular}
\vspace{\captionvspace}
\caption{
\textbf{Our results in contexts of other token-efficient video transformers on SomethingSomething-V2~\cite{goyal2017something}.}
For each method, we report numbers of tokens with respect to its corresponding baseline, the baseline accuracy, the proposed-model accuracy, and show the absolute performance change ($\Delta$) w.r.t the baseline.
Top block: exiting end-to-end methods.
We quote the numbers of STTS~\cite{wang2021efficient} and AdaFocus~\cite{wang2022adafocusv3} from their original publications.
Bottom block: our object-guided models.
Our model achieves better \emph{token}-efficiency compared to end-to-end methods with predicted boxes from detectors,
and can be further improved with ground truth boxes.
Note we do not claim better \emph{computation}-efficiency as our models require additional detection inputs.
}
\lbltbl{sota-efficiency}
\end{table}

\noindent\textbf{Token-efficient transformers.}
Finally, we put our results in the context of existing token-efficient transformers for videos.
We list perform of recent work STTS~\cite{wang2021efficient} and AdaFocus~\cite{wang2022adafocusv3},
both of which adaptively select video regions for action recognition on SomethingSomeghing-V2~\cite{goyal2017something}.
Note we are not strictly comparable to them in terms of accuracy-efficiency trade-off for two reasons.
First, our model requires additional object inputs, and overall costs more computes when taking into account the detector.
Also, different methods are developed on different video architectures, and we use performance change w.r.t. the baseline versus the remaining pixels as an evaluation proxy.
\reftbl{sota-efficiency} shows the results.
With bounding boxes from object detectors, our result of retaining baseline performance at $40\%$ tokens and is in a favorable position compared to STTS and AdaFocus.
Our results with ground truth boxes further highlight the advantages of using objects.

\vspace{-2mm}
\section{Discussion}

Our work is a first step towards a compact, object-based video processing framework.
First, we proposed an object-guided token sampling module that could drop background regions at an aggressive ratio.
Second, we proposed an object-aware attention module that utilizes object-token relation and improves action classification accuracy with minor additional computation.
Our overall framework improves both the token-efficiency and classification accuracy on popular action recognition benchmarks.

Despite promising preliminary results, 
there are several under-explored components in our framework.
First, it is still an open problem which object detector best suits the action recognition benchmark.
In our work, we found domain-specific detectors trained on human-action interaction datasets~\cite{materzynska2020something,shan2020understanding} worked the best, and switching to a stronger detection backbone~\cite{li2022exploring} or a more general open-world detector~\cite{zhou2022detecting} does not help.
It is unclear how object detection metrics (\ie the mAP on most datasets) correlate to the ability to help action recognition.
Second, running a detector requires substantial computation, particularly since modern detectors operate on high-resolution images.
As a result, our framework yet does not speed up the actual inference pipeline, and we leave integrating light-weighted object detectors within video frameworks as an exciting future direction.
We hope our work opens exciting avenues in the interaction between object detection and video recognition to future video architectures more efficient and accurate.

{\small
\bibliographystyle{ieee_fullname}
\bibliography{egbib}

\begin{thebibliography}{10}\itemsep=-1pt

\bibitem{arnab2021vivit}
Anurag Arnab, Mostafa Dehghani, Georg Heigold, Chen Sun, Mario Lu{\v{c}}i{\'c},
  and Cordelia Schmid.
\newblock Vivit: A video vision transformer.
\newblock In {\em ICCV}, 2021.

\bibitem{ba2016layer}
Jimmy~Lei Ba, Jamie~Ryan Kiros, and Geoffrey~E Hinton.
\newblock Layer normalization.
\newblock {\em arXiv:1607.06450}, 2016.

\bibitem{bertasius2021space}
Gedas Bertasius, Heng Wang, and Lorenzo Torresani.
\newblock Is space-time attention all you need for video understanding?
\newblock In {\em ICML}, 2021.

\bibitem{bewley2016simple}
Alex Bewley, Zongyuan Ge, Lionel Ott, Fabio Ramos, and Ben Upcroft.
\newblock Simple online and realtime tracking.
\newblock In {\em ICIP}, 2016.

\bibitem{bolya2022token}
Daniel Bolya, Cheng-Yang Fu, Xiaoliang Dai, Peizhao Zhang, Christoph
  Feichtenhofer, and Judy Hoffman.
\newblock Token merging: Your vit but faster.
\newblock {\em ICLR}, 2023.

\bibitem{jax2018github}
James Bradbury, Roy Frostig, Peter Hawkins, Matthew~James Johnson, Chris Leary,
  Dougal Maclaurin, George Necula, Adam Paszke, Jake Vander{P}las, Skye
  Wanderman-{M}ilne, and Qiao Zhang.
\newblock {JAX}: composable transformations of {P}ython+{N}um{P}y programs,
  2018.

\bibitem{cheng2022masked}
Bowen Cheng, Ishan Misra, Alexander~G Schwing, Alexander Kirillov, and Rohit
  Girdhar.
\newblock Masked-attention mask transformer for universal image segmentation.
\newblock In {\em CVPR}, 2022.

\bibitem{damen2020epic}
Dima Damen, Hazel Doughty, Giovanni~Maria Farinella, Sanja Fidler, Antonino
  Furnari, Evangelos Kazakos, Davide Moltisanti, Jonathan Munro, Toby Perrett,
  Will Price, et~al.
\newblock The epic-kitchens dataset: Collection, challenges and baselines.
\newblock {\em TPAMI}, 2020.

\bibitem{dehghani2022efficiency}
Mostafa Dehghani, Anurag Arnab, Lucas Beyer, Ashish Vaswani, and Yi Tay.
\newblock The efficiency misnomer.
\newblock In {\em ICLR}, 2022.

\bibitem{dehghani2021scenic}
Mostafa Dehghani, Alexey Gritsenko, Anurag Arnab, Matthias Minderer, and Yi
  Tay.
\newblock Scenic: A jax library for computer vision research and beyond.
\newblock In {\em CVPR}, 2022.

\bibitem{dosovitskiy2020image}
Alexey Dosovitskiy, Lucas Beyer, Alexander Kolesnikov, Dirk Weissenborn,
  Xiaohua Zhai, Thomas Unterthiner, Mostafa Dehghani, Matthias Minderer, Georg
  Heigold, Sylvain Gelly, et~al.
\newblock An image is worth 16x16 words: Transformers for image recognition at
  scale.
\newblock {\em ICLR}, 2021.

\bibitem{elsayed2022savi}
Gamaleldin~F Elsayed, Aravindh Mahendran, Sjoerd van Steenkiste, Klaus Greff,
  Michael~C Mozer, and Thomas Kipf.
\newblock Savi++: Towards end-to-end object-centric learning from real-world
  videos.
\newblock {\em NeurIPS}, 2022.

\bibitem{fan2021multiscale}
Haoqi Fan, Bo Xiong, Karttikeya Mangalam, Yanghao Li, Zhicheng Yan, Jitendra
  Malik, and Christoph Feichtenhofer.
\newblock Multiscale vision transformers.
\newblock In {\em ICCV}, 2021.

\bibitem{fayyaz2021ats}
Mohsen Fayyaz, Soroush~Abbasi Kouhpayegani, Farnoush~Rezaei Jafari, Eric
  Sommerlade, Hamid Reza~Vaezi Joze, Hamed Pirsiavash, and Juergen Gall.
\newblock Ats: Adaptive token sampling for efficient vision transformers.
\newblock {\em ECCV}, 2022.

\bibitem{feichtenhofer2022masked}
Christoph Feichtenhofer, Haoqi Fan, Yanghao Li, and Kaiming He.
\newblock Masked autoencoders as spatiotemporal learners.
\newblock {\em NeurIPS}, 2022.

\bibitem{feichtenhofer2019slowfast}
Christoph Feichtenhofer, Haoqi Fan, Jitendra Malik, and Kaiming He.
\newblock Slowfast networks for video recognition.
\newblock In {\em ICCV}, 2019.

\bibitem{goyal2017something}
Raghav Goyal, Samira Ebrahimi~Kahou, Vincent Michalski, Joanna Materzynska,
  Susanne Westphal, Heuna Kim, Valentin Haenel, Ingo Fruend, Peter Yianilos,
  Moritz Mueller-Freitag, et~al.
\newblock The" something something" video database for learning and evaluating
  visual common sense.
\newblock In {\em ICCV}, 2017.

\bibitem{grill2005visual}
Kalanit Grill-Spector and Nancy Kanwisher.
\newblock Visual recognition: As soon as you know it is there, you know what it
  is.
\newblock {\em Psychological Science}, 2005.

\bibitem{gupta2007objects}
Abhinav Gupta and Larry~S Davis.
\newblock Objects in action: An approach for combining action understanding and
  object perception.
\newblock In {\em CVPR}, 2007.

\bibitem{he2017mask}
Kaiming He, Georgia Gkioxari, Piotr Doll{\'a}r, and Ross Girshick.
\newblock Mask r-cnn.
\newblock In {\em ICCV}, 2017.

\bibitem{herzig2022object}
Roei Herzig, Elad Ben-Avraham, Karttikeya Mangalam, Amir Bar, Gal Chechik, Anna
  Rohrbach, Trevor Darrell, and Amir Globerson.
\newblock Object-region video transformers.
\newblock In {\em CVPR}, 2022.

\bibitem{huang2016deep}
Gao Huang, Yu Sun, Zhuang Liu, Daniel Sedra, and Kilian~Q Weinberger.
\newblock Deep networks with stochastic depth.
\newblock In {\em ECCV}, 2016.

\bibitem{jang2016categorical}
Eric Jang, Shixiang Gu, and Ben Poole.
\newblock Categorical reparameterization with gumbel-softmax.
\newblock {\em ICLR}, 2017.

\bibitem{kay2017kinetics}
Will Kay, Joao Carreira, Karen Simonyan, Brian Zhang, Chloe Hillier, Sudheendra
  Vijayanarasimhan, Fabio Viola, Tim Green, Trevor Back, Paul Natsev, et~al.
\newblock The kinetics human action video dataset.
\newblock {\em arXiv:1705.06950}, 2017.

\bibitem{kingma2015adam}
Diederik~P Kingma and Jimmy Ba.
\newblock Adam: A method for stochastic optimization. iclr. 2015.
\newblock {\em ICLR}, 2015.

\bibitem{kong2022spvit}
Zhenglun Kong, Peiyan Dong, Xiaolong Ma, Xin Meng, Wei Niu, Mengshu Sun, Xuan
  Shen, Geng Yuan, Bin Ren, Hao Tang, et~al.
\newblock Spvit: Enabling faster vision transformers via latency-aware soft
  token pruning.
\newblock In {\em ECCV}, 2022.

\bibitem{li2022exploring}
Yanghao Li, Hanzi Mao, Ross Girshick, and Kaiming He.
\newblock Exploring plain vision transformer backbones for object detection.
\newblock {\em ECCV}, 2022.

\bibitem{li2021improved}
Yanghao Li, Chao-Yuan Wu, Haoqi Fan, Karttikeya Mangalam, Bo Xiong, Jitendra
  Malik, and Christoph Feichtenhofer.
\newblock Mvitv2: Improved multiscale vision transformers for classification
  and detection.
\newblock In {\em CVPR}, 2022.

\bibitem{lin2019tsm}
Ji Lin, Chuang Gan, and Song Han.
\newblock Tsm: Temporal shift module for efficient video understanding.
\newblock In {\em ICCV}, 2019.

\bibitem{liu2021video}
Ze Liu, Jia Ning, Yue Cao, Yixuan Wei, Zheng Zhang, Stephen Lin, and Han Hu.
\newblock Video swin transformer.
\newblock {\em CVPR}, 2022.

\bibitem{locatello2020object}
Francesco Locatello, Dirk Weissenborn, Thomas Unterthiner, Aravindh Mahendran,
  Georg Heigold, Jakob Uszkoreit, Alexey Dosovitskiy, and Thomas Kipf.
\newblock Object-centric learning with slot attention.
\newblock {\em NeurIPS}, 2020.

\bibitem{materzynska2020something}
Joanna Materzynska, Tete Xiao, Roei Herzig, Huijuan Xu, Xiaolong Wang, and
  Trevor Darrell.
\newblock Something-else: Compositional action recognition with
  spatial-temporal interaction networks.
\newblock In {\em CVPR}, 2020.

\bibitem{meng2022adavit}
Lingchen Meng, Hengduo Li, Bor-Chun Chen, Shiyi Lan, Zuxuan Wu, Yu-Gang Jiang,
  and Ser-Nam Lim.
\newblock Adavit: Adaptive vision transformers for efficient image recognition.
\newblock In {\em CVPR}, 2022.

\bibitem{patrick2021keeping}
Mandela Patrick, Dylan Campbell, Yuki Asano, Ishan Misra, Florian Metze,
  Christoph Feichtenhofer, Andrea Vedaldi, and Jo{\~a}o~F Henriques.
\newblock Keeping your eye on the ball: Trajectory attention in video
  transformers.
\newblock {\em NeurIPS}, 2021.

\bibitem{radevski2021revisiting}
Gorjan Radevski, Marie-Francine Moens, and Tinne Tuytelaars.
\newblock Revisiting spatio-temporal layouts for compositional action
  recognition.
\newblock {\em BMVC}, 2021.

\bibitem{rao2021dynamicvit}
Yongming Rao, Wenliang Zhao, Benlin Liu, Jiwen Lu, Jie Zhou, and Cho-Jui Hsieh.
\newblock Dynamicvit: Efficient vision transformers with dynamic token
  sparsification.
\newblock {\em NeurIPS}, 2021.

\bibitem{ren2015faster}
Shaoqing Ren, Kaiming He, Ross Girshick, and Jian Sun.
\newblock Faster r-cnn: Towards real-time object detection with region proposal
  networks.
\newblock {\em NIPS}, 2015.

\bibitem{ryoo2021tokenlearner}
Michael~S Ryoo, AJ Piergiovanni, Anurag Arnab, Mostafa Dehghani, and Anelia
  Angelova.
\newblock Tokenlearner: What can 8 learned tokens do for images and videos?
\newblock {\em NeurIPS}, 2021.

\bibitem{shamsian2020learning}
Aviv Shamsian, Ofri Kleinfeld, Amir Globerson, and Gal Chechik.
\newblock Learning object permanence from video.
\newblock In {\em ECCV}, 2020.

\bibitem{shan2020understanding}
Dandan Shan, Jiaqi Geng, Michelle Shu, and David~F Fouhey.
\newblock Understanding human hands in contact at internet scale.
\newblock In {\em CVPR}, 2020.

\bibitem{szegedy2016rethinking}
Christian Szegedy, Vincent Vanhoucke, Sergey Ioffe, Jon Shlens, and Zbigniew
  Wojna.
\newblock Rethinking the inception architecture for computer vision.
\newblock In {\em CVPR}, 2016.

\bibitem{tenenbaum2011grow}
Joshua~B Tenenbaum, Charles Kemp, Thomas~L Griffiths, and Noah~D Goodman.
\newblock How to grow a mind: Statistics, structure, and abstraction.
\newblock {\em science}, 2011.

\bibitem{tong2022videomae}
Zhan Tong, Yibing Song, Jue Wang, and Limin Wang.
\newblock Videomae: Masked autoencoders are data-efficient learners for
  self-supervised video pre-training.
\newblock {\em NeurIPS}, 2022.

\bibitem{vaswani2017attention}
Ashish Vaswani, Noam Shazeer, Niki Parmar, Jakob Uszkoreit, Llion Jones,
  Aidan~N Gomez, {\L}ukasz Kaiser, and Illia Polosukhin.
\newblock Attention is all you need.
\newblock {\em NeurIPS}, 2017.

\bibitem{wang2021efficient}
Junke Wang, Xitong Yang, Hengduo Li, Zuxuan Wu, and Yu-Gang Jiang.
\newblock Efficient video transformers with spatial-temporal token selection.
\newblock {\em ECCV}, 2022.

\bibitem{wang2018temporal}
Limin Wang, Yuanjun Xiong, Zhe Wang, Yu Qiao, Dahua Lin, Xiaoou Tang, and Luc
  Van~Gool.
\newblock Temporal segment networks for action recognition in videos.
\newblock {\em TPAMI}, 2018.

\bibitem{wang2018videos}
Xiaolong Wang and Abhinav Gupta.
\newblock Videos as space-time region graphs.
\newblock In {\em ECCV}, 2018.

\bibitem{wang2021adaptive}
Yulin Wang, Zhaoxi Chen, Haojun Jiang, Shiji Song, Yizeng Han, and Gao Huang.
\newblock Adaptive focus for efficient video recognition.
\newblock In {\em ICCV}, 2021.

\bibitem{wang2022adafocus}
Yulin Wang, Yang Yue, Yuanze Lin, Haojun Jiang, Zihang Lai, Victor Kulikov,
  Nikita Orlov, Humphrey Shi, and Gao Huang.
\newblock Adafocus v2: End-to-end training of spatial dynamic networks for
  video recognition.
\newblock In {\em CVPR}, 2022.

\bibitem{wang2022adafocusv3}
Yulin Wang, Yang Yue, Xinhong Xu, Ali Hassani, Victor Kulikov, Nikita Orlov,
  Shiji Song, Humphrey Shi, and Gao Huang.
\newblock Adafocusv3: On unified spatial-temporal dynamic video recognition.
\newblock In {\em ECCV}, 2022.

\bibitem{wu2019long}
Chao-Yuan Wu, Christoph Feichtenhofer, Haoqi Fan, Kaiming He, Philipp
  Krahenbuhl, and Ross Girshick.
\newblock Long-term feature banks for detailed video understanding.
\newblock In {\em CVPR}, 2019.

\bibitem{wu2021towards}
Chao-Yuan Wu and Philipp Krahenbuhl.
\newblock Towards long-form video understanding.
\newblock In {\em CVPR}, 2021.

\bibitem{wu2018compressed}
Chao-Yuan Wu, Manzil Zaheer, Hexiang Hu, R Manmatha, Alexander~J Smola, and
  Philipp Kr{\"a}henb{\"u}hl.
\newblock Compressed video action recognition.
\newblock In {\em CVPR}, 2018.

\bibitem{wu2019detectron2}
Yuxin Wu, Alexander Kirillov, Francisco Massa, Wan-Yen Lo, and Ross Girshick.
\newblock Detectron2.
\newblock \url{https://github.com/facebookresearch/detectron2}, 2019.

\bibitem{yan2022multiview}
Shen Yan, Xuehan Xiong, Anurag Arnab, Zhichao Lu, Mi Zhang, Chen Sun, and
  Cordelia Schmid.
\newblock Multiview transformers for video recognition.
\newblock In {\em CVPR}, 2022.

\bibitem{yin2022vit}
Hongxu Yin, Arash Vahdat, Jose~M Alvarez, Arun Mallya, Jan Kautz, and Pavlo
  Molchanov.
\newblock A-vit: Adaptive tokens for efficient vision transformer.
\newblock In {\em CVPR}, 2022.

\bibitem{zhang2022object}
Chuhan Zhang, Ankush Gupta, and Andrew Zisserman.
\newblock Is an object-centric video representation beneficial for transfer?
\newblock {\em ACCV}, 2022.

\bibitem{zhang2017mixup}
Hongyi Zhang, Moustapha Cisse, Yann~N Dauphin, and David Lopez-Paz.
\newblock mixup: Beyond empirical risk minimization.
\newblock {\em ICLR}, 2018.

\bibitem{zhou2022detecting}
Xingyi Zhou, Rohit Girdhar, Armand Joulin, Phillip Kr{\"a}henb{\"u}hl, and
  Ishan Misra.
\newblock Detecting twenty-thousand classes using image-level supervision.
\newblock {\em ECCV}, 2022.

\bibitem{zhou2019objects}
Xingyi Zhou, Dequan Wang, and Philipp Kr{\"a}henb{\"u}hl.
\newblock Objects as points.
\newblock {\em arXiv:1904.07850}, 2019.

\end{thebibliography}
}

\end{document}